\documentclass[10pt,twocolumn,letterpaper]{article}

\usepackage{iccv}
\usepackage{times}
\usepackage{epsfig}
\usepackage{graphicx}
\usepackage{amsmath}
\usepackage{amssymb}
\usepackage{booktabs}
\usepackage{colortbl}
\usepackage{multirow}
\usepackage{multicol}
\usepackage{verbatim}

\usepackage[pagebackref,breaklinks,colorlinks]{hyperref}
\iccvfinalcopy 

\def\iccvPaperID{2796} 

\ificcvfinal\fi

\begin{document}
\title{Unified Data-Free Compression: Pruning and Quantization without Fine-Tuning}
\author{Shipeng Bai\thanks{Equal contribution.}
~ ~Jun Chen\footnotemark[1]
~ ~Xintian Shen 
~ ~Yixuan Qian
~ ~Yong Liu\thanks{Corresponding author.} \\
\normalsize College of Control Science and Engineering, Zhejiang University \\
{\tt\small [shipengbai, junc, 22132133, 22260066]@zju.edu.cn, yongliu@iipc.zju.edu.cn} 
}

\maketitle
\ificcvfinal\thispagestyle{empty}\fi

\begin{abstract}
Structured pruning and quantization are promising approaches for reducing the inference time and memory footprint of neural networks. 
However, most existing methods require the original training dataset to fine-tune the model. This not only brings heavy resource consumption but also is not possible for applications with sensitive or proprietary data due to privacy and security concerns. Therefore, a few data-free methods are proposed to address this problem, but they perform data-free pruning and quantization separately, which does not explore the complementarity of pruning and quantization.
In this paper, we propose a novel framework named Unified Data-Free Compression(UDFC), which performs pruning and quantization simultaneously without any data and fine-tuning process. Specifically, UDFC starts with the assumption that the partial information of a damaged(e.g., pruned or quantized) channel can be preserved by a linear combination of other channels, and then derives the reconstruction form from the assumption to restore the information loss due to compression. 
Finally, we formulate the reconstruction error between the original network and its compressed network, and theoretically deduce the closed-form solution.
We evaluate the UDFC on the large-scale image classification task and obtain significant improvements over various network architectures and compression methods. For example, we achieve a 20.54\% accuracy improvement on ImageNet dataset compared to SOTA method with 30\% pruning ratio and 6-bit quantization on ResNet-34.
Code will be available at \href{https://github.com/Dtudy/UDFC}{here}.
\end{abstract}

\begin{figure*}[t]
  \centering
   \includegraphics[width=\linewidth]{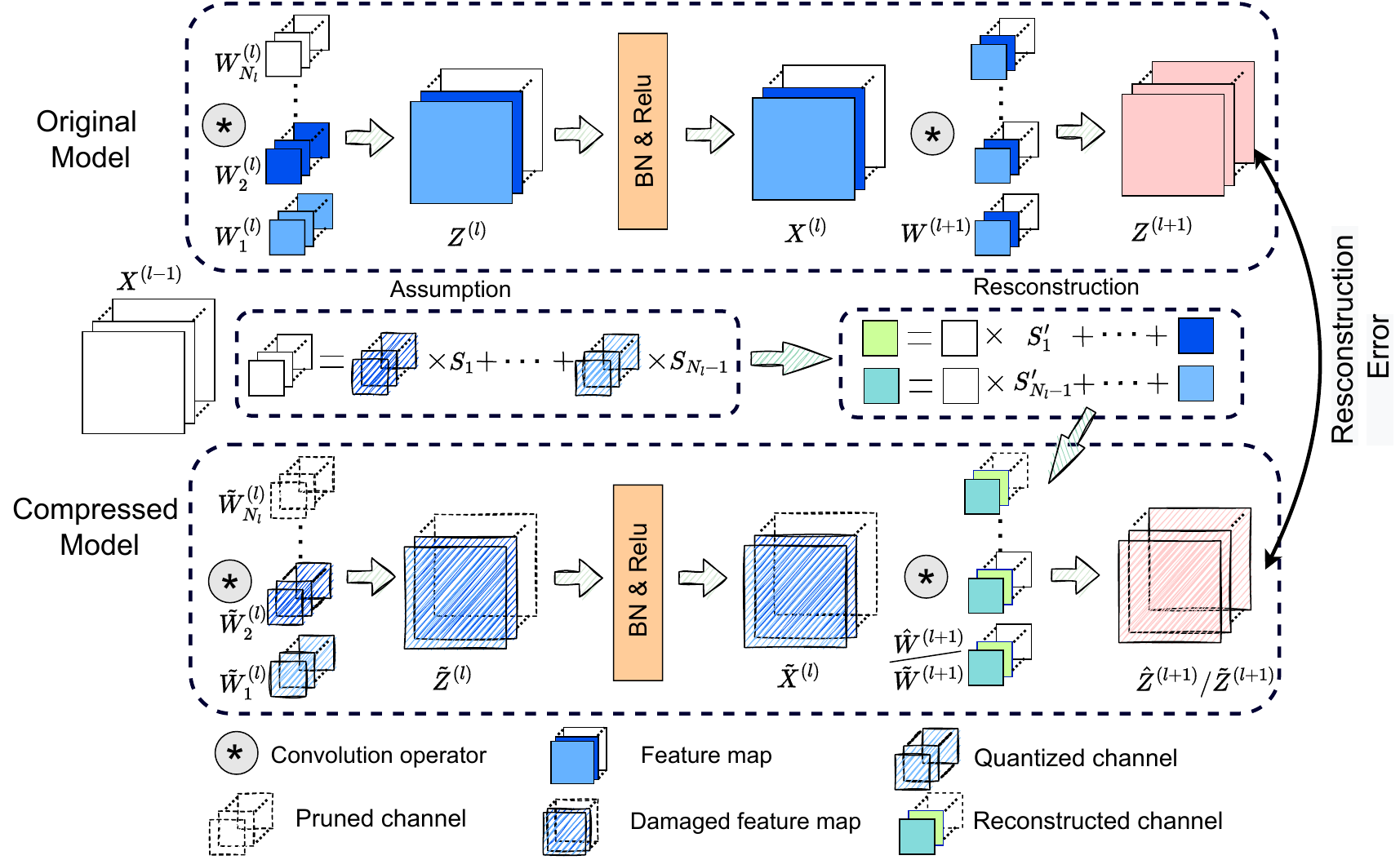}

   \caption{The general overview of UDFC, which performs the pruning and quantization simultaneously. $S$ is the scale factor and $S \neq S'$ . See Section \ref{sec: Formualtion of Reconstruction Error} for details of $S$. After the output channels of $l$-$th$ layer are pruned or quantized, our goal is to maintain the feature map $Z^{(l+1)}$ of $(l+1)$-$th$ layer. We first deduce the reconstruction form based on our assumption and then reconstruct the input channels of $(l+1)$-$th$ layer to restore the information loss caused by compression of $l$-$th$ layer. Finally, we formulate the reconstruction error between the feature map $Z^{(l+1)}$ and $\hat{Z}^{(l+1)}$/ $\Tilde{Z}^{(l+1)}$.}
   
   \label{fig:mianfig}
   
\end{figure*}

\section{Introduction}
\label{sec:intro}
Model compression is the most common way to reduce the memory footprint and computational costs of the model, and it mainly includes two methods: pruning\cite{Li2017PruningFF,Liu2019RethinkingTV,chen3} and quantization\cite{Gersho1991VectorQA,Jacob2018QuantizationAT,Wei2000FastTS,chen2020learning,chen2023data}. 
Among the pruning domain, structured pruning\cite{Wang2020StructuredPO,Sakai2022StructuredPF} is more actively studied than unstructured pruning\cite{Laurent2020RevisitingLM,Rumi2020AcceleratingSC} since it eliminates the whole channel or even the layer of the model while not requiring any special hardware or libraries for acceleration. Under such conditions, we also focus our attention on structured pruning in this paper.
Quantization methods attempt to reduce the precision of the parameters and/or activations from 32-bit floating point to low-precision representations. Thus the storage requirement for the model can be diminished substantially, as well as the power consumption.

Although the existing compression methods achieve a satisfactory compression ratio with a reasonable final performance, most of them require the original training data for a tedious fine-tuning process. The fine-tuning process is not only data-dependent but also computationally expensive, while users may not have access to sufficient or complete data in many real-world applications, such as medical data and user data. Therefore, data-free compression methods are proposed, which don't require any real data and fine-tuning process. 
For instance, Data-free parameter pruning\cite{srinivas2015data} first introduces the data-independent technique to remove the redundant neurons, and Neuron Merging\cite{kim2020neuron} extends the data-free method from fully connected layers to convolutional layers. Meanwhile, there exist some methods using the synthetic samples to perform the fine-tuning process, such as Dream\cite{yin2020dreaming}. 
In the field of quantization, recent works propose post-training quantization methods\cite{nagel2019data,cai2020zeroq,zhang2021diversifying,li2022dual,choi2021qimera, intraq} that use the synthetic data to replace the real data for quantization and achieve the SOTA results. For instance, ZeroQ \cite{cai2020zeroq} uses the distilled data that matches the statistics of batch normalization layers to perform post-training quantization. DSG\cite{zhang2021diversifying} proposes a novel Diverse Sample Generation scheme to enhance the diversity of synthetic samples, resulting in better performance.

However, some problems still hinder the deployment of data-free compression. On the one hand, the latest data-free quantization approaches focus on improving the quality of synthetic samples rather than releasing the quantization from its dependence on data. In this case, generating the synthetic samples introduces extra computational costs. On the other hand, current approaches perform data-free pruning and quantization separately, which does not explore the complementarity of weight pruning and quantization.

In this paper, we propose a novel joint compression framework named Unified Data-Free Compression(UDFC), which overcomes abovementioned issues without any original/synthetic data and fine-tuning process, as shown in Figure \ref{fig:mianfig}. Our contributions can be summarized as follows: 

\begin{itemize}
\item We propose the assumption that the partial information of a damaged(e.g., pruned or quantized) channel can be preserved by a linear combination of other channels. Based on this assumption, we derive that the information loss caused by pruning and quantization of the $l$-$th$ layer can be restored by reconstructing the channels of the $(l+1)$-$th$ layer. The assumption and reconstruction form are described in Section \ref{Layer-wise Reconstruction of Pruning}.

\item Based on the reconstruction form, we formulate the reconstruction error between the original network and its compressed network. The reconstruction error is described in Section \ref{tex:recon error}.

\item Based on the reconstruction error, we prove that reconstruction error can be minimized and theoretically deduce the closed form solution in Section \ref{tex:solution}. Furthermore, extensive experiments on CIFAR-10\cite{Krizhevsky2009LearningML} and ImageNet\cite{russakovsky2015imagenet} with various popular architectures demonstrate the effectiveness and generality of UDFC. For example, UDFC on VGG-16 yields around 70\% FLOPS reduction and 28× memory footprint reduction, with only a 0.4\% drop in accuracy compared to the uncompressed baseline on CIFAR-10 dataset.
\end{itemize}

\section{Related Work}
\subsection{Model Compression}
Researchers have proposed various methods to accelerate the model inference, mainly including network pruning\cite{han2015learning,liu2017learning} and network quantization\cite{Gupta2015DeepLW}. 
The early pruning methods concentrate around unstructured pruning, which removes single parameters from networks\cite{molchanov2017variational}.
These approaches, though theoretically interesting, are more difficult to implement within current hardware and software settings. Therefore, much recent work has focused on structured pruning\cite{anwar2017structured}, where network channels can be removed, and the models can be practically compressed and accelerated. 
Weight quantization refers to the process of discretizing the range of weight values so that each weight can be represented using fewer bits. \cite{Gupta2015DeepLW} first quantizes the network weights to reduce the model size by grouping the weights using k-means. \cite{hubara2016binarized,rastegari2016xnor} then introduce the binary network, in which weights are quantized to 1-bit. 
However, these aforementioned methods require access to data for fine-tuning to recover the performance. Fine-Tuning is often not possible for applications with sensitive or proprietary data due to privacy and security concerns. Therefore, we focus on pruning and quantization without any data and fine-tuning process in this paper.


\subsection{Data-Free Pruning}

Some pruning methods attempt to eliminate the dependency on the entire dataset and expensive fine-tuning process. \cite{luo2017thinet} formally establishes channel pruning as an optimization problem and solves this problem without using the entire dataset. \cite{mussay2019data} introduces a novel pruning algorithm, which can be interpreted as preserving the total flow of synaptic strengths through the network at initialization subject to a sparsity constraint. Meanwhile, there is a branch of data-free pruning methods \cite{tang2021data,tang2020reborn,yin2020dreaming} that still fine-tune the pruned model with limited or synthetically generated data. Although the above approaches propose effective methods for channel or neuron selection, several epochs of fine-tuning process and some training data are unavoidable to enable adequate recovery of the pruned network.

In fact, there are only two methods to prune the model without any data and fine-tuning process. Data-free parameter pruning\cite{srinivas2015data} shows how similar neurons are redundant and proposes a systematic way to remove them. Then Neuron Merging\cite{kim2020neuron} extends the data-free method from fully connected layers to convolutional layers based on the assumption that a pruned kernel can be replaced by another similar kernel.

\subsection{Data-Free Quantization}
Data Free Quantization \cite{nagel2019data} suffers a non-negligible performance drop when quantized to 6-bit or lower bit-width. Therefore, more recent studies employ generator architectures similar to GAN \cite{goodfellow2020generative} to generate synthetic samples that replace the original data. Such as, ZeroQ \cite{cai2020zeroq} generates samples that match the real-data statistics stored in the full-precision batch normalization layer to perform the post-training quantization resulting in better performance. IntraQ\cite{intraq} propose a local object reinforcement that locates the target 
objects at different scales and positions of synthetic images, aiming to enhance the intra-class heterogeneity in synthetic images. DSG\cite{zhang2021diversifying} slackens the batch normalization matching constraint and assigns different attention to specific layers for different samples to ensure diverse sample generation. 
However, using the generated samples to improve the accuracy of quantized models is time-consuming and complex. In this paper, we do not use any data to quantize the network.

\section{Formualtion of Reconstruction Error}
\label{sec: Formualtion of Reconstruction Error}

In this section, we first illustrate how to reconstruct the channels based on our assumption after pruning and quantization, and then mathematically formulate the reconstruction error. 

\subsection{Background Knowledge}
\paragraph{CNN architecture.}
Assuming that a CNN model with $L$ layers, we use $N_{l-1}$ and $N_{l}$ to represent the number of input channels and the output channels for the $l$-$th$ convolution layer. The $l$-$th$ convolution layer transforms the input feature map $X^{(l-1)} \in \mathbb{R}^{N_{l-1} \times H_{l-1} \times W_{l-1}}$ into the output feature map $Z^{(l)} \in \mathbb{R}^{N_{l} \times H_{l} \times W_{l}}$. The convolution weights of the $l$-$th$ layer are denoted as $W^{(l)} \in \mathbb{R}^{N_{l} \times N_{l-1} \times K \times K}$. Note that $K$ is the kernel size of each channel, and $H \times W$ is the corresponding feature map size. Therefore,

\begin{equation}
  {Z}^{(l)} = {X}^{(l-1)} \circledast {W}^{(l)},
  \label{eq:Z_L} 
\end{equation}
where $\circledast$ denotes the convolution operator. For CNN architectures, the convolution operation is widely followed by a batch normalization(BN) procedure and an activation function, thus the activation feature map ${X}^{(l+1)}$ can be formulated as:

\begin{equation}
  {X}^{(l)} = \Theta(\emph{B}({Z}^{(l)})) = \Theta(\frac{\gamma (Z^{(l)} - \mu)}{\sigma} + \beta ),
  \label{eq:BN and activation}
\end{equation}
in which $\emph{B}(\cdot)$ is the BN procedure and $\Theta(\cdot)$ is the activation function. $\gamma,\mu,\sigma$ and $\beta$ are the variables of BN.
\paragraph{Pruning criterion.}  
In channel pruning, most methods follow a selecting strategy, i.e., selecting some original channels via the $l_2$-norm\cite{Li2017PruningFF} of weight and scaling factors\cite{Liu2017LearningEC} in BN layer. In general, pruning criterion tends to be closely related to model performance. In this paper, we do not focus on proposing a complex criterion but on restoring the performance of networks that are pruned in a simple criterion such as $l_1$-norm and $l_2$-norm.

\paragraph{Uniform quantization.}
Quantization converts the floating-point parameters $W$ in the pretrained full-precision model to low-precision fixed-point values $\Tilde{W}$, thus reducing the model complexity. Uniform quantization \cite{zhou2016dorefa} is the simplest and most hardware-friendly method, which is defined as: 

\begin{equation}
\Tilde{W} = \frac{2}{2^k-1} round[(2^k-1)(\frac{W}{2max\vert W \vert}+ \frac{1}{2})]-1,
  \label{eq:quanization}
\end{equation}
where $k$ is the quantization bit-width. In this case, we use uniform quantization to preform the quantization process in this paper.

\subsection{Layer-wise Reconstruction }\label{Layer-wise Reconstruction of Pruning}
\paragraph{Assumption.}\label{assumption}
In model compression, the performance of the compressed network is usually worse than original network. To improve the performance of compressed network without any data and fine-tuning process, an ideal idea is to preserve the information of these damaged channels(e.g., pruned channel or quantized channel). We assume that the partial information of the damaged channels can be preserved by a linear combination of other channels. For clarity, we describe the assumptions about pruning and quantization separately. Suppose that convolution weight $W^{(l)}$ is pruned to its damaged versions $\hat{W}^{(l)} \in \mathbb{R}^{\hat{N}_{l} \times N_{l-1} \times K \times K}$, where $\hat{N}_{l}$ is the number of unpruned channels. The Assumption of pruning can be formulated as follows:
{\setlength\abovedisplayskip{7pt}
\setlength\belowdisplayskip{7pt}
\begin{equation}
  W_{j}^{(l)} \thickapprox \sum_{i=1}^{\hat{N}_{l}} \hat{s}_{i}\times W_{i}^{(l)}, \quad
  \forall j \in [\hat{N}_{l} , {N}_{l}],
   i \in [1 , \hat{N}_{l}]
  \label{eq:assumption_pruned}
\end{equation}}where $\hat{s}$ is a scale factor that measures the degree of association of the $i$-$th$ channel with the $j$-$th$ channel under the pruning. We prove that there always exists $\hat{s_i}$ minimizing the MSE error ($\Vert W_{j}^{(l)} - \sum_{i=1}^{\hat{N}_{l}} \hat{s}_{i}\times W_{i}^{(l)} \Vert_2^2$) of Eq.\ref{eq:assumption_pruned} in Section \ref{tex:solution}.

Suppose that the $m$-$th$ channel of $l$-$th$ layer is quantized to its damaged versions $\Tilde{W}_m^{(l)}$, the assumption of quantization can be formulated as:
{\setlength\abovedisplayskip{15pt}
\setlength\belowdisplayskip{7pt}
\begin{equation}
  \Tilde{W}_{m}^{(l)} \thickapprox  \Tilde{s}_{m}\times W_{m}^{(l)},\quad \forall m \in [1, N_l]
  \label{eq:assumption_quantization}
\end{equation}}where $\Tilde{s}$ is a scale factor that measures the degree of association of the $m$-$th$ channel with its quantized version under the quantization. We prove that there always exists $\Tilde{s_i}$ minimizing the MSE error ($\Vert \Tilde{W}_{m}^{(l)} -  \Tilde{s}_{m}\times W_{m}^{(l)} \Vert_2^2$) of Eq.\ref{eq:assumption_quantization} in Section \ref{tex:solution}.

\vspace{-15pt}
\paragraph{Reconstruction after pruning.}
Our goal is to maintain the output feature map of the $(l+1)$-$th$ layer while pruning the channels of the $l$-$th$ layer. For brevity, we prune only one channel in the $l$-$th$ layer to illustrate how the channels of $(l+1)$-$th$ layer are reconstructed, which can easily be generalized to multiple channels. Without loss of generality, the $j$-$th$ channel of the $l$-$th$ layer is to be pruned.


As shown in Figure \ref{fig:mianfig}, after the $j$-$th$ output channel of the $l$-$th$ layer is pruned, the output feature map $Z_j^{(l)}$ is subsequently deleted. Based on Eq.\ref{eq:Z_L} and Eq.\ref{eq:assumption_pruned}, we can deduce that the pruned output feature map $Z_j^{(l)}$ can be replaced by a linear combination of other undamaged feature maps:

\begin{equation}
\begin{aligned}
   Z_j^{(l)} &= X^{(l-1)} \circledast W_j^{(l)} \thickapprox X^{(l-1)} \circledast \sum_{i=1,i\neq j}^{{N_l}} \hat{s}_{i}\times W_{i}^{(l)} \\&= \sum_{i=1,i \neq j}^{{N_l}}\hat{s}_{i} \times Z_i^{(l)},
   \quad
\end{aligned}
 \label{eq:deduce}
\end{equation}
When only considering the BN layer, we have $X^{(l)}=\emph{B}(Z^{(l)})$. Based on Eq.\ref{eq:deduce}, the $k$-$th$ channel of output feature map $Z^{(l+1)}$ can be represented as:
{\setlength\abovedisplayskip{3pt}
\setlength\belowdisplayskip{1pt}
\begin{equation}
\begin{aligned}
   Z_k^{(l+1)} &= \sum_{i=1}^{{N_l}} X_{i}^{(l)} \circledast W_{k,i}^{(l+1)} \\
               &\thickapprox \sum_{i=1,i \neq j}^{{N_l}} \emph{B}(Z_{i}^{(l)}) \circledast (W_{k,i}^{(l+1)}+ \hat{s}_i \times W_{k,j}^{(l+1)}),\\ 
\end{aligned}
 \label{eq:reconstruction}
\end{equation}}
(More details in \textbf{Appendix A}.)

\noindent in which $(W_{k,i}^{(l+1)} + s_{i} \times W_{k,j}^{(l+1)})$ is a reconstructed filter. In this way, we can preserve the information of pruned channels in the $l$-$th$ layer by adding its corresponding pruned channel to each of the other channels in the next layer. According to the Eq.\ref{eq:reconstruction}, we reconstruct the channels of the $(l+1)$-$th$ layer to restore the information loss caused by pruning the $l$-$th$ layer in the following form:

\begin{equation}
\begin{aligned}
   \hat{Z}_k^{(l+1)} &= \sum_{i=1,i\neq j}^{{N_l}} X_{i}^{(l)} \circledast (W_{k,i}^{(l+1)} + \hat{s}_{i} \times W_{k,j}^{(l+1)}),
\end{aligned}
 \label{eq:compensate}
\end{equation}where $\hat{Z}_k^{(l+1)}$ represents the reconstructed version after pruning.
\paragraph{Reconstruction after quantization.}\label{Layer-wise Reconstruction of Quantization}
 The most significant difference between pruning and quantization is whether the channel exists or not. Quantized channels use the low bit-width to save the weights instead of discarding it away. In this case, we compensate for the information loss by adding a scale factor to its corresponding channels on the next layer. For simplicity, let $\Tilde{W}_m^{(l)}$ denotes the weight of $m$-$th$ channel of $l$-$th$ layer after quantization. Based on Eq.\ref{eq:Z_L} and Eq.\ref{eq:assumption_quantization}, we can deduce the reconstruction version of $\Tilde{Z}_k^{(l+1)}$ after quantization, which can be expressed as: 

\begin{equation}
\begin{aligned}
   \Tilde{Z}_k^{(l+1)} &= \sum_{i=1,i\neq m}^{{N_l}} X_{i}^{(l)} \circledast W_{k,i}^{(l+1)} + \Tilde{X}_{m}^{(l)} \circledast (\Tilde{s}_m \times W_{k,m}^{(l+1)}),
\end{aligned}
 \label{eq:quantization compensate}
\end{equation}
where $\Tilde{X}_{m}^{(l)}$ denotes the damaged version of ${X}_{m}^{(l)}$ after quantification.

\subsection{Reconstruction Error}
\label{tex:recon error}
However, the above analyses are all under the assumption, and the reconstruction error is inevitable in fact. After restoring information loss caused by the compression in the $l$-$th$ layer, we measure the reconstruction error using the feature map ${Z}_k^{(l+1)}$ of $(l+1)$-$th$ layer before and after compression. 
\vspace{-10pt}
\paragraph{Pruning error.}
After pruning, the difference $e_{p}$ between ${Z}_k^{(l+1)}$ and $\hat{Z}_k^{(l+1)}$ can be expressed as: 

\begin{equation}
\begin{aligned}
e_p &= {Z_k}^{(l+1)} - \hat{Z}_k^{(l+1)}  \\
         &=\{ \frac{\gamma_j}{\sigma_j} \{X^{(l-1)}\circledast (W_j^{(l)}- \sum_{i=1,i \neq j}^{{N_l}} \hat{s}_i \frac{\gamma_i \sigma_j}{\sigma_i \gamma_j} W_i^{(l)})\} +  \\
         &\quad (\beta_j - \frac{\gamma_j \mu_j}{\sigma_j}) - (\sum_{i=1,i \neq j}^{{N_l}} \hat{s}_i (\beta_i - \frac{\gamma_i \mu_i}{\sigma_i}))\} \circledast W_{k,j}^{(l+1)}
  \end{aligned}
  \label{eq:error}
\end{equation}
(More details in \textbf{Appendix B}.)

\paragraph{Influence of Activation Function.}
The Relu activation function is widely used in CNN architectures. Since we cannot obtain the feature map after the activation function in a data-free way, we qualitatively analyze the effects of the Relu function on our pruning error $e_p$. In this case, the difference $e_{p}$ can be re-expressed as:

\begin{equation}
\begin{aligned}
 e_p &= \Theta(\emph{B}(Z_j^{(l)})) -  \sum_{i=1,i\neq j}^{{N_l}} \hat{s}_{i} \times \Theta(\emph{B}(Z_i^{(l)}))\\
      &\leqslant \frac{1}{2} (A + \vert A \vert),\\
 \end{aligned}
\label{eq:activation}
\end{equation}
(More details in \textbf{Appendix C}.)

\noindent where $A = \emph{B}(Z_j^{(l)}) - \sum_{i=1,i\neq j}^{{N_l}} \hat{s}_{i} \times \emph{B}(Z_i^{(l)})$ and we omit the $W_{k,j}^{(l+1)}$ as it doesn't change with pruning. The term $\frac{1}{2}(A + \vert A \vert)$ determine the upper boundary of $e_p $ and the form of $(\emph{B}(Z_j^{(l)}) - \sum_{i=1,i\neq j}^{{N_l}} \hat{s}_{i} \times \emph{B}(Z_i^{(l)})\circledast W_{k,j}^{(l+1)}$ is the same as Eq.\ref{eq:error}, so the difference $e_p$ of pruning we obtained is equal 
whether the Relu activation function is considered or not.

Note that $X^{(l-1)}$ and $W_{k,j}^{(l+1)}$ are not changed with pruning. Therefore, we define the reconstruction error $\ell_{p}$ of pruning as:

\begin{equation}
\begin{aligned}
 \ell_{p} &= \Vert W_j^{(l)}- \sum_{i=1,i \neq j}^{{N_l}}\hat{s}_i \frac{\gamma_i \sigma_j}{\sigma_i \gamma_j} W_i^{(l)}\Vert_2^2 \\
 &+ \alpha_1 \Vert (\beta_j - \frac{\gamma_j \mu_j}{\sigma_j}) - \sum_{i=1,i \neq j}^{{N_l}} \hat{s}_i (\beta_i - \frac{\gamma_i \mu_i}{\sigma_i}) \Vert_2^2,
 \end{aligned}
\label{eq:re_error}
\end{equation}
in which, we introduce a hyperparameter $\alpha_1$ to adjust the proportion of different parts.

\paragraph{Quantization error.} 
After quantization, the difference $e_{q}$ of ${Z}_k^{(l+1)}$ can be expressed as: 

\begin{equation}
\begin{aligned}
  e_q &= {Z_k}^{(l+1)} - \Tilde{Z}_k^{(l+1)}  \\
       &= \{(\frac{\gamma_m W_m^{(l)}}{\sigma_m} - \Tilde{s}_m\frac{{\gamma}_m \Tilde{W}_m^{(l)}}{{\sigma}_m})\circledast X^{(l-1)} + \Tilde{s}_m \frac{{\gamma}_m {\mu}_m}{ {\sigma}_m}  \\ 
       & \quad - \frac{\gamma_m \mu_m}{\sigma_m} + \beta_m - \Tilde{s}_m {\beta}_m\}\circledast W_{k,m}^{(l+1)}, \\
  \end{aligned}
\end{equation}
(More details in \textbf{Appendix D}.)

Same as pruning, $X^{(l-1)}$ and $W_{k,m}^{(l+1)}$ are not changed with quantization, while the activation function does not influence the form of the reconstruction error. Therefore, we define the reconstruction error $\ell_{q}$ of quantization as:

\begin{equation}
\begin{aligned}
  \ell_{q} &= \Vert \frac{\gamma_m W_m^{(l)}}{\sigma_m} - \Tilde{s}_m\frac{{\gamma}_m \Tilde{W}_m^{(l)}}{{\sigma}_m}) \Vert_2^2 + \\ 
  & \quad \alpha_2 \Vert (\beta_m - \frac{\gamma_m \mu_m}{\sigma_m}) - \Tilde{s}_m ({\beta}_m - \frac{{\gamma}_m {\mu}_m}{ {\sigma}_m}) \Vert_2^2
  \end{aligned}
\end{equation} in which, we introduce a hyperparameter $\alpha_2$ to adjust the proportion of different parts.
\paragraph{Reconstruction error}
Previously, we analyzed the errors caused by pruning and quantization separately. When pruning and quantization are performed simultaneously, the reconstruction error $\ell_{re}$ can be expressed as:
{\setlength\abovedisplayskip{7pt}
\setlength\belowdisplayskip{7pt}
\begin{equation}
\begin{aligned}
  \ell_{re} = \ell_{p} + \ell_{q}
  \end{aligned}
\end{equation}}

\section{Solutions for Reconstruction Error} \label{tex:solution}
In this section, we prove the existence of the optimal solution $s$ by minimizing the reconstruction error. The $j$-$th$ channel is pruned and the $m$-$th$ channel is quantized in $l$-$th$ layer, and we get the reconstruction error $\ell_{re}$:
\begin{equation}
\begin{aligned}
 \ell_{re} &= \Vert W_j^{(l)}-\hat{s}_i \sum_{i=1,i \neq j}^{{N_l}} \frac{\gamma_i \sigma_j}{\sigma_i \gamma_j} W_i^{(l)}\Vert_2^2 \\
 &+ \alpha_1 \Vert (\beta_j - \frac{\gamma_j \mu_j}{\sigma_j}) - (\sum_{i=1,i \neq j}^{{N_l}} \hat{s}_i (\beta_i - \frac{\gamma_i \mu_i}{\sigma_i})) \Vert_2^2\\
 & +  \Vert \frac{\gamma_m W_m^{(l)}}{\sigma_m} - \Tilde{s}_m\frac{{\gamma}_m \Tilde{W}_m^{(l)}}{{\sigma}_m}) \Vert_2^2  \\ 
  & + \alpha_2 \Vert (\beta_m - \frac{\gamma_m \mu_m}{\sigma_m}) - \Tilde{s}_m ({\beta}_m - \frac{{\gamma}_m {\mu}_m}{ {\sigma}_m}) \Vert_2^2,
 \end{aligned}
\label{eq:ove}
\end{equation}
For simplicity, we have:
{\setlength\belowdisplayskip{7pt}
\begin{equation}
\begin{aligned}
   \begin{cases}
    
    \textbf{G}_i = \frac{\gamma_i \sigma_j}{\sigma_i \gamma_j}\textbf{W}_i^{(l)}, \quad \textbf{V} = {W}_j^{(l)}, \\
    \textbf{Q} = [\textbf{G}_1,\cdots ,\textbf{G}_{j-1},\textbf{G}_{j+1},\cdots, \textbf{G}_{N_l}], \\
    \textbf{K}_i = \beta_i - \frac{\gamma_i \mu_i}{\sigma_i}, \\
    \textbf{P} = [\textbf{K}_1,\cdots ,\textbf{K}_{j-1},\textbf{K}_{j+1},\cdots, \textbf{K}_{N_l}], \\
    \textbf{R}_i = \frac{\gamma_i \textbf{W}_i^{(l)}}{\sigma_i},\\
    \hat{\textbf{s}} = [\hat{s}_1,\cdots,\hat{s}_{j-1},\hat{s}_{j+1},\cdots,\hat{s}_{N_l}], \\
    \end{cases}
 \end{aligned}
\label{eq:simplicity}
\end{equation}}
where $\textbf{V}$ is the vectorized ${W}_i^{(l)}$, $\textbf{K}_i$ is the vectorized $\beta_i - \frac{\gamma_i \mu_i}{\sigma_i}$ and $\textbf{R}_i$ is the vectorized $\frac{\gamma_i \textbf{W}_i^{(l)}}{\sigma_i}$. Then the loss can be simplified as follows:

{\setlength\abovedisplayskip{1pt}
\setlength\belowdisplayskip{7pt}
\begin{equation}
\begin{aligned}
     \ell_{re} &= (\textbf{V}-\hat{\textbf{s}}\textbf{Q})^T(\textbf{V}-\hat{\textbf{s}}\textbf{Q}) + \alpha_1 (\textbf{K}_j - \hat{\textbf{s}}\textbf{P})^T(\textbf{K}_j - \hat{\textbf{s}}\textbf{P}) \\ 
 &+ (\textbf{R}_m-\Tilde{s}_m{\textbf{R}_m})^T(\textbf{V}-\Tilde{s}_m{\textbf{R}_m}) \\ &+ \alpha_2(\textbf{K}_m - \Tilde{s}_m\textbf{K}_m)^T(\textbf{K}_m - \Tilde{s}_m{\textbf{K}_m}) 
 \end{aligned}
\label{eq:simplified}
\end{equation}}
The first and second derivative of the $\hat{\textbf{s}}$ is:
{\setlength\abovedisplayskip{7pt}
\setlength\belowdisplayskip{7pt}
\begin{equation}
\begin{aligned}
 \frac {\partial \ell_{re}} {\partial \hat{\textbf{s}}} &= -2\textbf{Q}^T\textbf{V}+2\hat{\textbf{s}}\textbf{Q}^T\textbf{Q} + \alpha_1 (-2\textbf{P}^T \textbf{K}_j+2\hat{\textbf{s}}\textbf{P}^T \textbf{P})\\
 \frac {\partial^{2} \ell_{re}}{\partial \hat{\textbf{s}}^{2}} &= 2\textbf{Q}^T\textbf{Q} + 2 \alpha_1 \textbf{P}^T \textbf{P} \\
 \end{aligned}
\label{eq:prove}
\end{equation}}
$\ell_{re}$ is a convex function and thus there exists an unique optimal solution $s$ such that $\frac {\partial \ell_{re}} {\partial \hat{\textbf{s}}} = 0$, so we get the optimal solution as follows:
{\setlength\abovedisplayskip{7pt}
\setlength\belowdisplayskip{7pt}
\begin{equation}
\begin{aligned}
\hat{\textbf{s}} = (\textbf{Q}^T\textbf{V} + \alpha_1 \textbf{P}^T \textbf{K}_j)(\textbf{Q}^T\textbf{Q} + \alpha_1 \textbf{P}^T\textbf{P})^{-1} \\
 \end{aligned}
\label{eq:hatxsolution}
\end{equation}}
Similarly, we get the optimal solution of $\Tilde{s}_m$:
{\setlength\abovedisplayskip{7pt}
\setlength\belowdisplayskip{7pt}
\begin{equation}
\begin{aligned}                                                                                
\Tilde{s}_m = ({\textbf{R}_m}^T\textbf{R}_m + \alpha_2 \textbf{K}_m^T \textbf{K}_m)({\textbf{R}_m}^T{\textbf{R}_m} + \alpha_2 \textbf{K}_m^T\textbf{K}_m)^{-1}
 \end{aligned}
\label{eq:tildexsolution}
\end{equation}}

{\setlength\abovedisplayskip{7pt}
\setlength\belowdisplayskip{7pt}
\begin{figure}[b]
  \centering
   \includegraphics[width=0.9\linewidth]{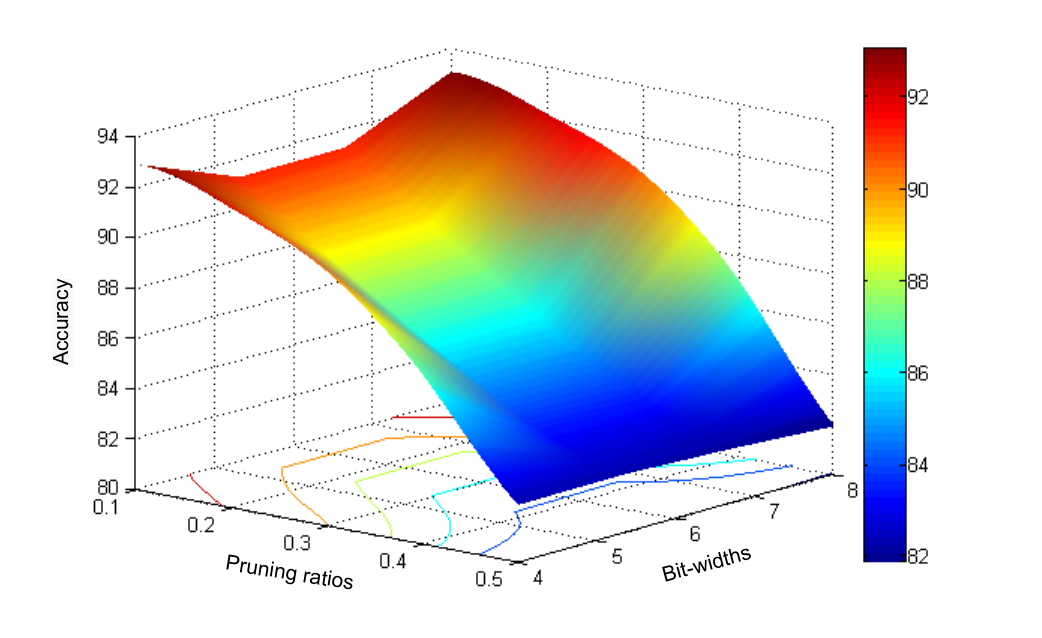}

   \caption{Comparison of the accuracy of ResNet-56 with different pruning ratios and bit widths on CIFAR-10 dataset.}
   \label{fig:paq}
\end{figure}}

It is worth noting that the MSE error of Eq.\ref{eq:assumption_pruned} and Eq.\ref{eq:assumption_quantization} are the main components of the reconstruction error. Therefore, the optimal solution s not only minimizes the reconstruction error but also satisfies the assumptions. 
\begin{figure}[t]
  \centering
  \setlength{\belowcaptionskip}{-3mm}
   \includegraphics[width=0.9\linewidth]{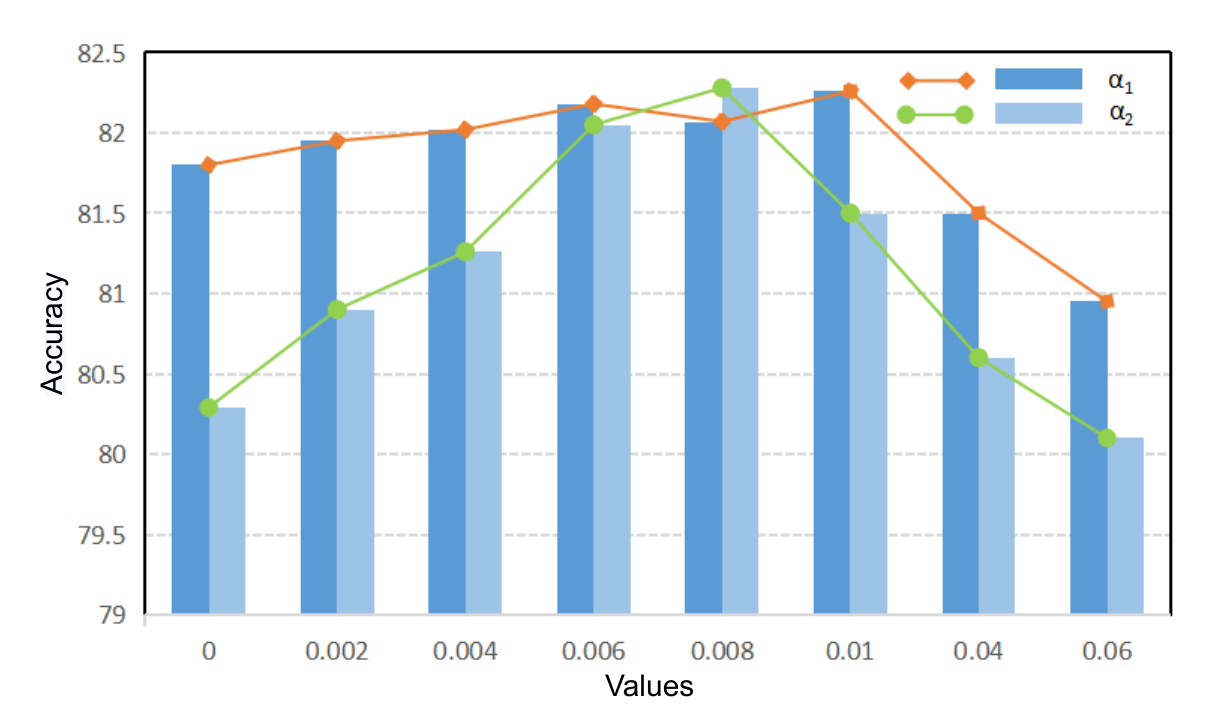}

 \caption{The accuracy comparison of different $\alpha$ values on ResNet-56. As the value $\alpha$ increases, the accuracy curve first rise and then fall.}
   \label{fig:hpy}
\end{figure}

\paragraph{Implementation of the scale factors.}
After getting the optimal scales, we replace the original convolutional layer with the reconsturction form, which are shown in Eq.\ref{eq:compensate} and Eq.\ref{eq:quantization compensate}. 
\begin{table}[b]
  \centering
   \caption{Quantization results on ImageNet dataset. 'No-D' denotes whether to use synthetic samples or calibration sets.}
   \vspace{5pt}
  \resizebox{0.5\textwidth}{!}{
  \begin{tabular}{l|lcccc}
    \toprule
    Model & Method &No-D & (W/A)Bit  & Size(MB) & Top-1($\%$) \\
    \midrule
    \multirow{3}{*}{ResNet-18}  & Baseline & -  & 32/32 &44.59& 71.47 \\
    & DFQ\cite{nagel2019data} &$\checkmark$  & 6/6 & 8.36 & 66.30 \\
    & DSG\cite{zhang2021diversifying}&$\times$ & 6/6 &8.36 & 70.46 \\
    & SQuant\cite{squant}&$\times $& 6/6 &8.36& 70.74 \\
    &    Ours &$\checkmark$  & 6/6  & 8.36& \textbf{72.76}\\
    & DDAQ\cite{li2022dual} &$\times$ & 4/4  &5.58& 58.44 \\
    & DSG\cite{zhang2021diversifying}&$\times$  & 4/4 &5.58& 34.53 \\
    &    Ours &$\checkmark$ & 4/4  & 5.58& \textbf{63.49} \\
    
    \hline
    \multirow{9}{*}{ResNet-50}& Baseline & - & 32/32 &97.49& 77.72 \\
    & ZeroQ\cite{cai2020zeroq}&$\times$  & 6/6  &18.46 & 75.56\\
    & DSG\cite{zhang2021diversifying}&$\times$ & 6/6  &18.46 &  76.07\\
    & SQuant\cite{squant}&$\times$ & 6/6  &18.46 & 77.05\\
    & Ours &$\checkmark$ & 6/6  & 18.46 & \textbf{77.57}\\
    & OMSE\cite{choukroun2019low}&$\checkmark$ & 4/32 &12.28  & 67.36\\
    & GDFQ\cite{xu2020generative}&$\times$  & 4/4 &12.28 & 55.65\\
    & SQuant\cite{squant}&$\times$ & 4/4  &12.28 &  70.80\\
    &  Ours &$\checkmark$ & 4/4 & 12.28 & \textbf{72.09}\\
    \hline
    \multirow{5}{*}{MobileNetV2}& Baseline&  -  & 32/32 & 13.37& 73.03\\
    & DFQ\cite{nagel2019data} &$\checkmark$ & 8/8 &3.34 & 71.20\\
    & DDAQ\cite{li2022dual}&$\times$  & 6/6 &2.50 & 70.30\\
    & ZeroQ\cite{cai2020zeroq}&$\times$  & 6/6 &2.50& 69.62 \\
     &   Ours &$\checkmark$  & 6/6 & 2.50 & \textbf{71.87}\\
     
     \hline
    \multirow{3}{*}{DenseNet}& Baseline &  -  & 32/32 &31.92& 74.36\\
    & OMSE\cite{choukroun2019low}&$\checkmark$  & 4/32 &6.00  & 64.40\\
     &   Ours & $\checkmark$ & 4/32& 6.00 &\textbf{70.15}\\
    \toprule 
 \end{tabular}}
 
  \label{tab:quantization}
\end{table}

\begin{table*}[t]
  \centering
    \caption{Results of VGG-16 and ResNet-56 on CIFAR-10 dataset. 'P-R' represents the pruning ratio. 'Ave-Im' denotes the accuracy improvement compared to Prune. 'W-bit' denotes the bit-width of the weights.}
    \vspace{5pt}
  \begin{tabular}{l|l|c|c|c|c|c}
    \toprule
    \multirow{11}{*}{\rotatebox{90}{VGG-16 (Acc.93.70)}} &
    \multirow{2}{*}{P-R} & \multirow{2}{*}{Method} & Criterion & \multirow{2}{*}{Ave-Im($\uparrow$)} & \multirow{2}{*}{W-bit} & \multirow{2}{*}{Size(MB)}  \\

    & & & $l_2$-norm \quad $l_1$-norm  &  & & \\
    \cline{2-7}
     &\multirow{3}{*}{60\%} & Prune &   89.14  \qquad    88.70  & 0 & 32 & 21.6  \\ 
    &         & NM    &   93.16 \qquad        93.16  & 4.24 & 32 & 21.6  \\
     &        & Ours  & \textbf{93.40 } \qquad \textbf{93.40} & \textbf{4.48}& \textbf{6}  & \textbf{4.04} \\
    \cline{2-7}
    & \multirow{3}{*}{70\%}& Prune &   35.83  \qquad         35.55  & 0& 32 & 16.4  \\ 
     &        & NM    &        65.77 \qquad          65.35 & 29.87 & 32 & 16.4  \\
     &        &  Ours  & \textbf{93.32} \qquad \textbf{93.20} & \textbf{57.31} & \textbf{6}  & \textbf{3.08} \\
    \cline{2-7}
    &\multirow{3}{*}{80\%} & Prune  &         18.15 \qquad   17.56     & 0 & 32 & 11.2  \\ 
     &        & NM    &         40.26 \qquad        39.49  & 22.02& 32 & 11.2  \\
     &        &  Ours  & \textbf{91.79} \qquad \textbf{91.26} & \textbf{73.67} & \textbf{6}  & \textbf{2.12} \\
    \toprule
    \toprule
    \multirow{9}{*}{\rotatebox{90}{ResNet-56 (Acc.93.88)}} 
    &\multirow{3}{*}{30\%} & Prune &  76.95 \qquad       74.46   & 0 &  32 & 2.4   \\
    &         & NM    &  85.22 \qquad       84.41   & 9.11 & 32 & 2.4 \\
    &   &  Ours  & \textbf{90.33} \qquad \textbf{90.28}& \textbf{14.6}  & \textbf{4}  & \textbf{0.30} \\
    \cline{2-7}
    &\multirow{3}{*}{40\%} & Prune & 46.44 \qquad 49.68 & 0 & 32 & 2.0 \\ 
    &         & NM    & 76.56 \qquad 77.89 & 24.16  & 32 & 2.0  \\
     &        &  Ours  & \textbf{86.99} \qquad \textbf{87.29} &  \textbf{39.08} & \textbf{4}  & \textbf{0.24} \\
    \cline{2-7}
    &\multirow{3}{*}{50\%} & Prune & 24.34 \qquad 25.58 & 0 & 32 & 2.15  \\ 
    &         & NM    & 56.18 \qquad 56.45 & 31.36 & 32  & 2.15  \\
    &         &  Ours  & \textbf{81.90} \qquad \textbf{81.60} & \textbf{56.79} & \textbf{4}  & \textbf{0.20} \\
    \bottomrule
  \end{tabular}

  \label{tab:CIFAR-10}
\end{table*}

\section{Experiments}\label{experiments}
In this section, we conduct experiments with several different widely-used neural networks for image classification task to evaluate the effectiveness of our data-free compression method that do not require any data and fine-tuning process. In all experiments, we quantize the weights of all model layers using uniform quantization and prune the channels with the simple pruning criterion $l_1$-norm and $l_2$-norm. 
In addition, we visualize the weights offset and loss landscape \cite{visualloss} to further illustrate the validity of UDFC, and the results are shown in \textbf{Appendix E}.

\subsection{Ablation Study}
Our proposed method consists of two compression techniques, quantization and pruning. Meanwhile, there exist hyperparameters $\alpha_1$ and $\alpha_2$ in the reconstruction error that impacts the compressed network performance. We perform the following ablation studies to evaluate the effects of different parts of our framework.


\vspace{-10pt}
\paragraph{Study on pruning ratio and bit-width.}
UDFC performs pruning and quantization, the appropriate variables(i.e., pruning ratio and bit-width) become critical to compression ratio and performance of compressed model. Therefore, we compress ResNet-56 with different pruning ratios and bit-widths on CIFAR-10\cite{lebedev2016fast} dataset to explore the optimal trade-off between pruning and quantization.

As shown in Figure \ref{fig:paq}, the model performance decreases as the pruning ratio gradually increases. Similarly, the model performance also decreases as the weight bit width decreases, but at 4-bit the accuracy does not drop but rises. This peculiar phenomenon indicates that our method can maximize the restoration of information when quantizing ResNet-56 with 4-bit. We do not present quantified results for lower bits(i.e., 3-bit and 2-bit) because their accuracy drops sharply. At lower bits quantization, the loss of information is so great that our method cannot effectively restore the information.

\vspace{-10pt}
\paragraph{Study on hyperparameters $\alpha_1$ and $\alpha_2$.}
To explore the effect of hyperparameters $\alpha_1$ and $\alpha_2$ on  compressed network, we prune and quantize ResNet-56 separately on CIFAR-10 dataset. During the pruning, we use different $\alpha_1$ values to prune 50\% channels of ResNet-56. During the quantization, we use different $\alpha_2$ values for 2-bit quantization of ResNet-56.

As shown in Figure \ref{fig:hpy}, when $\alpha_1$ increases from 0 to 0.01, the final performance of the pruned model increase steadily. However, when $\alpha_1$ is set to 0.04, the accuracy suffers a huge drop. The curve of $\alpha_2$ is similar to that of $\alpha_1$, with the maximum performance at 0.008. This phenomenon confirms that the hyperparameters we introduced have improved the performance of the compressed model to some extent. 

\begin{table*}[t]
  \centering
    \caption{Results of ResNet-101 and ResNet-34 on ImageNet dataset. 'P-R' represents the pruning ration. 'Ave-Im' denotes the accuracy improvement compared to Prune. 'W-bit' denotes the bit-width of weights. }
      \vspace{5pt}
  \begin{tabular}{l|l|c|c|c|c|c|c}

    \toprule
    \multirow{11}{*}{\rotatebox{90}{ResNet-101 (Acc.77.31\%)}} &
    \multirow{2}{*}{P-R} & 
    \multirow{2}{*}{Method} & Criterion & \multirow{2}{*}{Ave-Im ($\uparrow$)} & \multirow{2}{*}{W-bit} & \multirow{2}{*}{Size(MB)} & \multirow{2}{*}{FLOPS(G)} \\

    & & & $l_2$-norm \quad $l_1$-norm  &  & & & \\
    \cline{2-8}
     &\multirow{3}{*}{10\%} & Prune &   69.10  \qquad        68.52  & 0 & 32 & 154.4 &6.84 \\ 
    &         & NM    &   72.46 \qquad        71.95  & 3.40 & 32 & 154.4& 6.84  \\
     &        &  Ours  & \textbf{74.69 } \qquad \textbf{74.61} & \textbf{5.84} & \textbf{6}  & \textbf{28.8} &\textbf{6.84} \\
    \cline{2-8}
    & \multirow{3}{*}{20\%}& Prune &   45.60  \qquad         44.45  & 0& 32 & 132.4 &6.08 \\ 
     &        & NM    &         62.41 \qquad          60.57 & 16.46 & 32 & 132.4 & 6.08 \\
     &        &  Ours  & \textbf{71.36} \qquad \textbf{71.00} & \textbf{26.16} & \textbf{6}  & \textbf{24.8} &\textbf{6.08} \\
    \cline{2-8}
    &\multirow{3}{*}{30\%} & Prune  & 10.10 \qquad 9.560 & 0 & 32 & 112.4 &5.3 \\ 
     &        & NM    &         38.44 \qquad        37.68  & 28.23& 32 & 112.4 & 5.3  \\
     &        &  Ours  & \textbf{65.76} \qquad \textbf{65.22} & \textbf{55.66} & \textbf{6}  & \textbf{21.2} &\textbf{5.3}\\
    \toprule
    \toprule
    
    \multirow{11}{*}{\rotatebox{90}{\qquad  ResNet-34 (Acc.73.27\%)}}&
    
    \multirow{3}{*}{10\%} & Prune &  63.51 \qquad  61.95   & 0 &  32 & 78.8  & 3.24\\
    &         & NM    &          67.10 \qquad        66.50   & 4.35 & 32 & 78.8& 3.24 \\
    &         &  Ours  & \textbf{69.86} \qquad \textbf{69.39}& \textbf{6.89} &  \textbf{6}  & \textbf{14.8} & \textbf{3.24} \\
    \cline{2-8}
    &\multirow{3}{*}{20\%} & Prune & 42.80 \qquad 40.62 & 0 & 32 & 70.0  & 2.88\\ 
    &         & NM    & 55.70 \qquad 54.20 & 13.24  & 32 & 70.0  &2.88\\
     &        &  Ours  & \textbf{65.44} \qquad \textbf{64.68}  & \textbf{23.35} & \textbf{6}  & \textbf{13.2}&\textbf{2.88}  \\
    \cline{2-8}
    &\multirow{3}{*}{30\%} & Prune & 16.80 \qquad 12.60 & 0 & 32 & 61.6  & 2.52\\ 
    &         & NM    & 39.40 \qquad 36.34 & 23.17 & 32  & 61.6  & 2.52\\
    &         &  Ours  & \textbf{59.25} \qquad \textbf{57.57} & \textbf{43.71} & \textbf{6}  & \textbf{11.6}&\textbf{2.52} \\

    
    \bottomrule
  \end{tabular}

  \label{tab:ImageNet}
\end{table*}

\subsection{Quantization}
We quantize ResNet-18\cite{he2016deep}, ResNet-50\cite{he2016deep}, MobileNetV2\cite{howard2017mobilenets} and DenseNet\cite{huang2017densely} on ImageNet\cite{russakovsky2015imagenet} dataset using the uniform quantization. 
In order to demonstrate the effectiveness of our method on quantization, we compare our method with DFQ\cite{nagel2019data}, OMSE\cite{choukroun2019low} and SQuant\cite{squant}, which do not require any data and fine-tuning process. In addition, we also compare our method with some Post Training Quantization(PTQ) methods including ZeroQ\cite{cai2020zeroq}, DDAQ\cite{li2022dual}, DSG\cite{zhang2021diversifying} and ZAQ\cite{liu2021zero}, which use the synthetic samples or calibration sets to improve the performance of quantized model.

Table \ref{tab:quantization} shows that our method has significant advantages compared to DFQ, OMSE and other PTQ methods for various models. For instance, when quantizing the weights of ResNet-18 with 6-bit, our method achieves 72.76$\%$ accuracy that is 6.46\% higher than DFQ and 1.9\% higher than DSG. Our method remains robust to low-bit quantization of the lightweight model MobileNetV2(71.87\%) and DenseNet(70.15\%). In addition, our method has a tremendous advantage in time consumption. ZeroQ takes 29 seconds to quantize ResNet50 on an 8 Tesla V100 GPUs, while UDFC only takes 2 seconds on a RTX 1080Ti GPU. 


\subsection{Unified Compression.}
In this section, we compress the ResNet-56 and VGG-16\cite{Simonyan2015VeryDC} on CIFAR-10\cite{Krizhevsky2009LearningML} dataset, ResNet-34 and ResNet-101 on ImageNet dataset to demonstrate the effectiveness of our method. Since no data-free method can perform both pruning and quantization simultaneously, we mainly compare our method with data-free pruning methods. In the field of pruning, our direct competitor is Neuron Merging(NM)\cite{kim2020neuron}, which is a one-to-one compensation method. Same as Neuron Merging, we do not perform any compensation after pruning as a way to obtain the baseline performance, called $Prune$.

\vspace{-10pt}
\paragraph{Experiments on CIFAR-10.}
For the CIFAR-10 dataset, we test UDFC on ResNet-56 and VGG-16 with different pruning rates: 30\%-80\%. In addition, we quantize the unpruned channels to 4-bit and 6-bit respectively, further reducing the memory footprint of parameters.

As shown in Table \ref{tab:CIFAR-10}, UDFC achieves state-of-the-art performance. For example, with about 28$\times$ parameters drop(0.53M) and 80\% FLOPS reduction, VGG-16 still has excellent classification accuracy (91.26\%), which is 51\% average accuracy higher than NM at a 80\% pruning ratio.
\vspace{-9pt}
\paragraph{Experiments on ImageNet.}
For the ImageNet dataset, we test UDFC on ResNet-34 and ResNet-101 with pruning rates: 10\%, 20\% and 30\%. In addition, we quantize the unpruned channels to 6-bit, further reducing the memory footprint of parameters. 

Table \ref{tab:ImageNet} shows that UDFC outperforms the previous method. 
By varying the ratio of pruning from 10\% to 30\%, the Ave-Im increases accordingly compared to NM and Prune. That means our method is more robust than one-to-one compensation.
For ResNet-101, we get a 55.66\% improvement in accuracy compared to Prune and a 27.23\% improvement compared to NM at a pruning ratio of 30\%. 
Meanwhile, the parameters are substantially reduced due to the quantization, so that not only do we achieve higher performance but also a lower memory footprint of parameters both in ResNet-34 and ResNet-101.

\section{Conclusion}
In this paper, we propose a unified data-free compression framework that performs pruning
and quantization simultaneously without any data and fine-tuning process. It starts with the assumption that the partial information of a damaged channel can be preserved by a linear combination of other channels and then gets a fresh approach from the assumption to restore the information loss caused by compression. Extensive experiments on benchmark datasets validate the effectiveness of our proposed method.

\section{Acknowledgement}
This work was supported by a Grant from The National Natural Science Foundation of China(No. U21A20484)
{\small
\bibliographystyle{iee}
\bibliography{egbib}
}

\end{document}